\documentclass[journal]{IEEEtran}

\usepackage{amsmath,amsfonts}
\usepackage{algorithmic}
\usepackage[ruled]{algorithm2e}
\usepackage{xcolor,soul,framed} %,caption

\colorlet{shadecolor}{yellow}
\usepackage{color,soul}
\usepackage[pdftex]{graphicx}
\usepackage{array}
\usepackage[hyphens]{url}
\usepackage{multirow}
\usepackage{amsmath}
\usepackage{threeparttable} % 引入包

\usepackage{subfigure}
\usepackage{graphicx}
\usepackage{multirow}
\usepackage{hyperref}
\usepackage{bbm}
\usepackage{bm}

\begin{document}

\title{LearningFlow: Automated Policy Learning Workflow for Urban Driving with\\ Large Language Models}

\author{
Zengqi Peng, Yubin Wang, Xu Han, Lei Zheng,
and Jun Ma
\thanks{Zengqi Peng, Yubin Wang, and Lei Zheng are with the Robotics and Autonomous Systems Thrust, The Hong Kong University of Science and Technology (Guangzhou), China (email: zpeng940@connect.hkust-gz.edu.cn; ywang575@connect.hkust-gz.edu.cn; lzheng135@connect.hkust-gz.edu.cn). }
\thanks{Xu Han is with the Data Science and Analytics Thrust, The Hong Kong University of Science and Technology (Guangzhou), China (email: xhanab@connect.ust.hk). }
\thanks{Jun Ma is with the Robotics and Autonomous Systems Thrust, The Hong Kong University of Science and Technology (Guangzhou), China, and also with the Department of Electronic and Computer Engineering, The Hong Kong University of Science and Technology, Hong Kong SAR, China
(e-mail: jun.ma@ust.hk).
}

}

\maketitle

\begin{abstract}
Recent advancements in reinforcement learning (RL) demonstrate the significant potential in autonomous driving. Despite this promise, challenges such as the manual design of reward functions and low sample efficiency in complex environments continue to impede the development of safe and effective driving policies.
To tackle these issues, we introduce LearningFlow, an innovative automated policy learning workflow tailored to urban driving. This framework leverages the collaboration of multiple large language model (LLM) agents throughout the RL training process. LearningFlow includes a curriculum sequence generation process and a reward generation process, which work in tandem to guide the RL policy by generating tailored training curricula and reward functions. Particularly, each process is supported by an analysis agent that evaluates training progress and provides critical insights to the generation agent.
Through the collaborative efforts of these LLM agents, LearningFlow automates policy learning across a series of complex driving tasks, and it significantly reduces the reliance on manual reward function design while enhancing sample efficiency. 
Comprehensive experiments are conducted in the high-fidelity CARLA simulator, along with comparisons with other existing methods, to demonstrate the efficacy of our proposed approach. The results demonstrate that LearningFlow excels in generating rewards and curricula. It also achieves superior performance and robust generalization across various driving tasks, as well as commendable adaptation to different RL algorithms.

\end{abstract}

\begin{IEEEkeywords}
Autonomous driving, large language model, automated policy learning, curriculum reinforcement learning.
\end{IEEEkeywords}

\section{Introduction}
 
With the advancement of artificial intelligence (AI) technologies, significant breakthroughs have been made in generative models, with large language models (LLMs) being one of the most prominent applications \cite{zhao2023survey,chang2024survey}. 
LLMs excel in understanding and generating text, while their integration with additional specialized modules enables multimodal capabilities, such as processing and generating images and videos. This inherent versatility demonstrates the immense potential for application across various fields \cite{cui2024survey}. 
Meanwhile, autonomous driving technology has made remarkable progress, becoming a focal point of research in AI and transportation \cite{cui2024survey,kiran2021deep,muhammad2020deep}. In general, urban driving scenarios are characterized by diverse road structures and task types, such as multi-lane overtaking, on-ramp merging, and intersection crossing. These driving scenarios demand frequent interactions with surrounding vehicles (SVs) exhibiting varying driving styles. The diversity of driving environments and the uncertainty of SV behaviors collectively present significant challenges to achieving a safe and efficient closed-loop urban driving system \cite{mozaffari2020deep,ettinger2021large}. Consequently, autonomous driving systems are required to enhance the ability to prevent potential collision risks and optimize task efficiency across different scenarios. From this perspective, urban autonomous driving requires robust interaction-aware decision-making and planning capabilities to safely interact with SVs while efficiently accomplishing various driving tasks.

\begin{figure*}[!htbp] %[htbp]
    \centering
    \includegraphics[trim=0 0cm 0 0.5cm, width=0.8\linewidth]{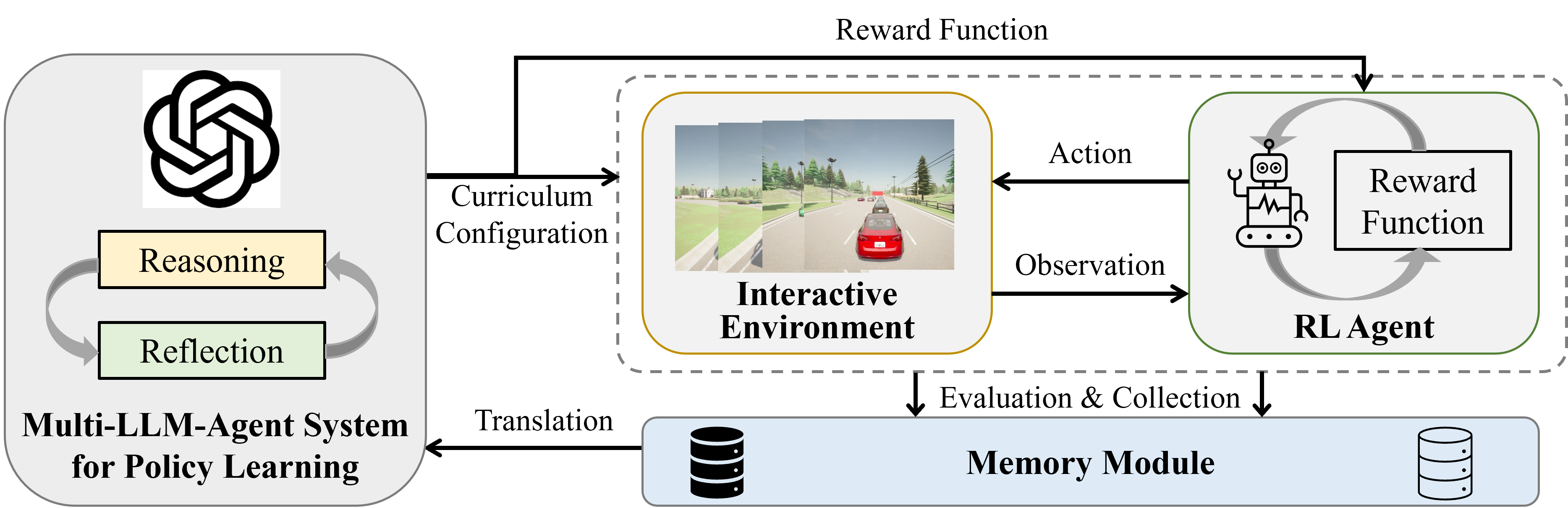}
   \caption{The proposed LLM-in-the-training-loop CRL training paradigm. The Multi-LLM-agent system generates reward functions and training curriculum sequences for the downstream RL policy through the collaboration of multiple LLM-based agents. The historical training data generated through interactions between the RL policy and the environment is stored in a memory module and then fed back to the Multi-LLM-agent system as reference information for subsequent generation steps.
   }
    \label{intro:paradigm}
\end{figure*}

Reinforcement learning (RL) has demonstrated significant potential for autonomous driving solutions. It enables policies to optimize decision-making through interactions with the environment and the feedback received from these interactions \cite{kiran2021deep,qiao2021behavior,wang2023chance,peng2024reward}. 
Despite the significant progress of RL in the autonomous driving community, it still encounters two major challenges. The first challenge lies in the design of reward functions, which RL relies on to guide agents in exploring the environment and improving policies. However, in real-world tasks like autonomous driving, reward signals are often sparse, and this significantly hinders efficient policy learning \cite{booth2023perils}. Reward shaping is a common approach to provide incremental learning signals to mitigate this issue \cite{hu2020learning}. Nevertheless, designing an appropriate reward function for complex autonomous driving tasks remains highly challenging. 
Traditional manual design methods are not only constrained by the subjective experience of the designer but also time-consuming and tedious \cite{sutton2018reinforcement,ma2023eureka,abouelazm2024review}. 
Due to the distinct characteristics of different tasks, the reward functions that are capable of effectively guiding RL policies to learn satisfactory behaviors could vary significantly. 
Furthermore, the reward function is typically fixed at the beginning of the policy training and cannot be adapted in real-time during the training process. 
Therefore, the manual design of reward functions often fails to encapsulate the nuanced behaviors needed and provide effective guidance for RL agents in dynamic urban driving scenarios, leading to poor policy convergence. 
The second challenge is the sample efficiency during online exploration. The complexities of urban driving tasks involve diverse road structures, varying task objectives, traffic densities, and the driving styles of SVs. Relying on RL policies to learn from samples collected through random exploration in the target task is highly inefficient, potentially requiring an enormous number of training episodes and resulting in unstable performance \cite{narvekar2020curriculum}. 
Curriculum learning (CL) is a structured approach inspired by human learning, which improves training efficiency by decomposing tasks in a progressive order from simple to complex \cite{bengio2009curriculum,soviany2022curriculum}. This approach ultimately enables the agent to acquire effective policies for complex tasks. 
While CL provides a promising framework for addressing training inefficiencies in RL for complex tasks, challenges remain in evaluating and dynamically switching between training curricula. Several automated CL algorithms have been proposed to assess training progress and quality, dynamically adjusting the training curriculum at appropriate stages \cite{graves2017automated,peng2024reward}. 
The training task is modeled as a multi-armed bandit problem, with a curriculum set of distinct difficulty levels to enable curriculum transitions. However, updating the multi-armed bandit algorithm requires incorporating new supervision signals, and parameter tuning is task-specific. This problem could lead to inefficiency when dealing with different scenarios.

To tackle the aforementioned challenges, this paper investigates the integration of curriculum RL (CRL) and LLMs to develop a policy training framework, LearningFlow, which aims to automate curriculum transition and reward design for urban driving tasks. 
As an emerging technology, LLMs demonstrate remarkable capabilities in problem reasoning and code generation, offering new routines to address the challenges of curriculum selection and reward design in CRL solutions of autonomous driving. 
By leveraging the vast knowledge base of LLMs, we aim to address the limitations of traditional RL approaches in autonomous driving, enhancing both the scalability and performance of RL-based driving policies. 
Therefore, this paper presents a novel automated policy training workflow, which consists of multiple LLM agents, for CRL in urban driving. 
The overview of the proposed training paradigm is shown in Fig. \ref{intro:paradigm}. Here, the RL policy is trained with the support of a multi-LLM-agent system, which iteratively generates and refines the training curricula and reward functions throughout the training process. 
To the best of our knowledge, this work is the first iterative and online framework that leverages LLMs for automated reward function optimization and curriculum transition in urban driving. 
The main contributions are listed as follows:

\vspace{-0.1cm}
\begin{itemize}
    \item A novel automated policy learning framework is proposed for training interaction-aware driving policies in complex autonomous driving tasks, which significantly improves driving safety and generalization ability under different traffic densities.

    \item A collaborative role-playing module involving multiple LLM agents is developed to iteratively generate training curricula and reward functions. This module effectively enhances the exploration of RL agents, thereby significantly improving the training efficiency of driving policies.

    \item  We demonstrate the effectiveness of the proposed approach in the high-fidelity simulator CARLA. The proposed method exhibits effective curriculum generation and reward generation capabilities. It achieves superior performance and commendable generalization ability across various driving tasks and RL algorithms.

\end{itemize}

The rest of the paper is structured as follows. Section II presents the related work. Section III introduces the problem formulation. Section IV illustrates the proposed framework for automated policy learning. Section V demonstrates the experimental results followed by pertinent analysis. Finally, Section VI summarizes the conclusion and discusses future works.

\section{Related Work}

\subsection{Reward Design for Deep Reinforcement Learning}

The reward function plays a critical role in guiding RL agents as they interact with the environment. A high-quality reward function can significantly enhance the performance of RL policies. Therefore, the design of reward functions, often referred to as reward engineering, plays a vital role in RL. Designing suitable reward functions for real-world tasks, such as autonomous driving and robotics, poses challenges due to the sparsity of rewards over long time horizons and the balance between exploration and exploitation. In existing studies, manual design is the most commonly used routine for constructing reward functions, 
followed by minor adjustments through trial and error \cite{peng2024reward,zhuang2023robot,al2023self}. This approach not only makes the effectiveness of the reward function dependent on expert experience but also renders the entire training process time-consuming and labor-intensive. Furthermore, when dealing with complex multi-task problems, relying solely on expert experience often fails to yield a satisfactory reward function. Multi-task RL techniques are introduced by sharing policies or designing task-specific policies \cite{kai2020multi,sodhani2021multi,liu2023multi}, which still do not consider the reward design problems. 

To address the challenges arising from reward design, inverse RL (IRL) is utilized to extract reward functions from collected data by observing the agent behaviors of interaction with the environment \cite{fu2017learning,arora2021survey}. 
Specifically, the deep IRL methods are adopted to infer the driving behaviors by distilling the learned reward model from collected expert demonstrations to score and evaluate trajectories in autonomous driving tasks \cite{phan2022driving,nan2023interaction}. 
However, due to the non-uniqueness of reward functions, the inferred reward function could not fully capture the true motivations of the agent. On the other hand, IRL relies on a large amount of expert data, and collecting high-quality data in complex and dynamic environments is challenging, if not impractical. This reward inference method typically involves complex optimization processes, resulting in heavy computational burdens, long training time, and limitations due to environmental modeling. Additionally, evolutionary algorithms have been introduced for reward shaping to evolve reward functions \cite{hu2020learning,devidze2022exploration}. 
With the advancement of foundation model technologies, the design of reward functions for RL tasks can be accomplished by providing relevant prompts to LLMs \cite{kwon2023reward,li2024auto}, which offers a promising solution to solve the challenges of reward design in RL.

\subsection{Training Efficiency for Deep Reinforcement Learning} 

RL policies improve performance by interacting with the environment to collect experience. However, low sample efficiency remains a significant challenge in applying RL to complex tasks. Directly employing random policies to gather sample information in intricate environments is highly inefficient, which can significantly prolong the time required for policy convergence or even lead to failure in achieving convergence. 
To tackle this issue, an environment model is introduced to generate virtual samples to improve the sample efficiency, thereby accelerating the training process \cite{guan2020centralized}. However, the effectiveness of the trained policy highly depends on the accuracy of the model. 

CL offers a promising solution to mitigate the above challenges \cite{bengio2009curriculum,song2021autonomous}. 
A stage-decaying CL approach is utilized to guide the policy learning of the RL agents \cite{peng2023CPPO}. 
Nevertheless, the predetermined manual scheduling of curriculum transitions heavily relies on expert knowledge, which limits the robustness and effectiveness of the training outcomes. 
To tackle the aforementioned challenges, various automated CL methods have been proposed \cite{qiao2018automatically,khaitan2022state}. 
However, these approaches are designed under the assumption that SVs do not respond to the behaviors of the ego vehicle (EV). Besides, the future trajectories of SVs are accessible to the EV. These simplifications could compromise driving safety and reduce the generalization capability of RL policies. 
In real-world scenarios, human drivers dynamically adapt their driving maneuvers based on the behaviors of SVs, which is critical for ensuring safe and efficient interactions on the road. 
Essentially, as most current studies do not comprehensively consider the interactive behaviors of SVs, these simplifications could limit the deployment of autonomous driving techniques in real-world scenarios.

\subsection{Large Language Model Applications}

With the rapid advancement of LLMs, their potential in various tasks has attracted significant attention \cite{cui2024survey}. Currently, LLMs are primarily utilized in two ways. The first one is the LLMs-in-the-task-loop solution, which utilizes LLMs for embodied inference. The second one is the LLMs-in-the-training-loop solution, which incorporates LLMs for policy learning.

\subsubsection{LLMs for Embodied Inference}
One of the most direct ways to apply LLMs technology to autonomous driving tasks is by embedding the LLMs as a decision-making or planning module within the autonomous driving system. Depending on the embedding method, there are primarily two approaches, serial LLMs solutions and parallel LLMs solutions \cite{wen2023dilu,xu2024drivegpt4,wang2023drivemlm}. For example, DiLu achieves decision-making based on common-sense knowledge by integrating an LLMs-based inference module into the autonomous driving system \cite{wen2023dilu}. 
However, this sequential structure limits the response speed of the autonomous driving system due to the inference latency of LLMs, posing challenges to meeting real-time requirements. 
To address this issue, DriveVLM-Dual connects a traditional end-to-end pipeline with an LLMs-based inference system in a parallel manner, forming a fast-slow system that alleviates the real-time performance challenges of LLMs-based autonomous driving solutions \cite{tian2024drivevlm}.

\subsubsection{LLMs for Policy Learning}
In addition to directly deploying LLMs within autonomous driving systems,  LLMs are also applied during the training phase \cite{li2024auto,yu2023language,hazra2024revolve,zhou2024context}. 
Recent research has explored the use of LLMs for designing reward functions in reinforcement learning and assisting in CL for task decomposition. A general RL training framework has been proposed to generate proxy reward functions by leveraging the in-context learning capabilities and prior knowledge of LLMs \cite{kwon2023reward}. Auto MC-Reward enhances learning efficiency by automatically designing dense reward functions through the introduction of three automated reward modules \cite{li2024auto}. Eureka is proposed as an LLM-driven human-level reward generation method for sequential decision-making tasks \cite{ma2023eureka, ma2024dreureka}. 
CurricuLLM is proposed to decompose complex robotic skills into a sequence of subtasks, facilitating the learning of intricate robotic control tasks \cite{ryu2024curricullm}. 
However, the entire training curriculum sequence is predetermined before training and the timing of curriculum transitions is neglected, which potentially hinders the sample efficiency. 
AutoReward leverages LLMs and CoT-based modules to achieve closed-loop reward generation for autonomous driving \cite{han2024autoreward}. 
However, this approach requires access to the internal code of the simulation environment, which could result in the leakage of environmental dynamics. 
It does not account for traffic density, requiring retraining for driving tasks with varying traffic conditions in the same driving environment. Additionally, AutoReward adjusts the reward function after the whole training cycle, which increases training time and could result in unnecessary computational costs.

\section{Problem Formulation}
\label{Section2}
\subsection{Problem Statement}

This study aims to tackle the challenge of developing safe, effective, and interaction-aware driving policies for various urban driving scenarios, such as multi-lane overtaking and on-ramp merging. The initial and target positions of both the EV and SVs are randomly generated, while ensuring adherence to traffic regulations. The number of SVs is also random, with interactive behaviors driven by different driving styles. In this context, the EV is required to complete various driving tasks within an environment that contains interactive SVs. Since different driving tasks have distinct characteristics and requirements, this presents a significant challenge in designing appropriate reward functions and suitable training curricula. 
The goal is to automatically generate reward functions and training curricula to train RL policies that infer decision sequences, guiding the EV to safely and efficiently complete driving tasks across different scenarios. 
Here, we assume that the EV can access the exact position and velocity information of SVs accurately, yet their goal tasks and driving intentions are unknown. These configurations inject a significant level of randomness into the driving scenarios, rendering the tasks challenging but close to real-world situations.

\subsection{Learning Environment}

In this work, the target tasks are formulated as a Markov Decision Process (MDP). 
Here, we represent the MDP as a tuple $\mathcal{E} = \langle \mathcal{S}, \mathcal{A}, \mathcal{P}, \mathcal{R}, \gamma \rangle$, with each element defined as follows:

\textbf{State space $\mathcal{S}$}: In this work, $\mathcal{S}$ includes kinematic features of driving vehicles within the observation range of the EV. The state matrix at time step $k$ is defined as shown below:
\begin{equation}
\begin{split}
\mathbf{S}_{k}=\left[\ \mathbf{s}_{k}^0\ \ \mathbf{s}_{k}^1\ ...\ \mathbf{s}_{k}^{N_{\text{sv}}^{\max}}\ \right]^T,
\end{split}
\label{state martix}
\end{equation}
where $N_{\text{sv}}^{\max}$ indicates the maximum number of SVs observed by the EV; $\mathbf{s}_{k}^0$ and $\mathbf{s}_{k}^i\ (i=1,2,...,N_{\text{sv}}^{\max})$ denote the state of the EV and the state of the $i$-th SV, respectively. 
In particular, $\mathbf{s}_{k}^i$ is defined as follows:
\begin{equation}
\begin{split}
\mathbf{s}_{k}^{i}={\left[\begin{array}{l l l l l l l}{x_{k}^{i}}&{y_{k}^{i}}&{v_{k}^{i}}&{\psi_{k}^{i}}\end{array}\right]}^{T},
\end{split}
\label{state_space}
\end{equation}
where $x_{k}^{i}$, $y_{k}^{i}$, $v_{k}^{i}$, $\psi_{k}^{i}$ are the X-axis and Y-axis coordinates, the speed, and the heading angle of the $i$-th vehicle, respectively. 

\textbf{Action space $\mathcal{A}$}: In this work, a multi-discrete action space consisting of three discrete sub-action spaces is utilized for the RL agent:
\begin{equation}
\begin{split}
\mathcal{A}=\left\{ A_{1},A_{2},A_{3} \right\},
\end{split}
\label{Action_space}
\end{equation}
where $A_{1}, A_{2}$, and $A_{3}$ denote the waypoint, reference velocity, and lane change sub-action spaces, respectively. Further details will be provided in Section \ref{method:RL executor}.

\textbf{State transition dynamics $\mathcal{P}(\mathbf{S}_{k+1}|\mathbf{S}_{k},a_{k})$}: $\mathcal{P}$ describes the transitions of the environmental state while satisfying the Markov property. It is implicitly determined by the external environment and remains inaccessible to the RL agent.

\textbf{Reward function $\mathcal{R}$}: The reward function plays a crucial role in RL. It reinforces the correct actions of the agent by providing rewards and penalizes incorrect actions, guiding the exploration of the agent within the environment. A well-designed reward function can significantly enhance the efficiency and performance of the training process. However, designing rewards manually for complex tasks remains challenging. In this work, we leverage the extensive knowledge base of LLMs to design and iteratively refine efficient reward functions for RL agents.

\textbf{Discount factor $\gamma$}: $\gamma \in (0,1)$ is utilized to discount future accumulated rewards.

\subsection{Curriculum Sequence Generation Problem}

Under the problem defined in Section III.A, we establish the following two-layer curriculum set. The first layer considers the traffic densities, while the second layer takes into account the motion modes of SVs. Specifically, the designed curriculum set consists of \( N_{td} \) subsets, each comprising \( N_{mm} \) elements. This two-layer curriculum set can be expressed as:

\begin{equation}
\begin{split}
\boldsymbol{\Omega}=\{\Omega_{i j} \mid i=0,1, \ldots, N_{\mathrm{td}}^{\max }, j=0,1, \ldots, N_{\mathrm{mm}}^{\max } \} ,
\end{split}
\label{2layer_curri_set}
\end{equation}
where $N_{\mathrm{td}}^{\max }$ and $N_{\mathrm{mm}}^{\max }$ denote the number of traffic density types and motion modes of SVs, respectively. 
In CRL, a sequence of training curricula is required to set up the environment for optimizing the RL policy, which can be expressed as follows: 
\begin{equation}
\{\Omega^*_{l m}\}=\arg \max \sum\nolimits \mathcal{R}, %\textbf{R},
\label{Equ:optimal_curri}
\end{equation}
where $\{\Omega^*_{l m}\}$ is the optimal curriculum sequence, and $\mathcal{R}$ is the reward of the RL agent. In this study, the curriculum sequence is formulated through the use of LLMs, which are employed for their robust generative abilities to guide the learning process.

\begin{figure*}[!t] %[htbp]
    \centering
    \includegraphics[trim=0 0cm 0 0cm, width=0.9\linewidth]{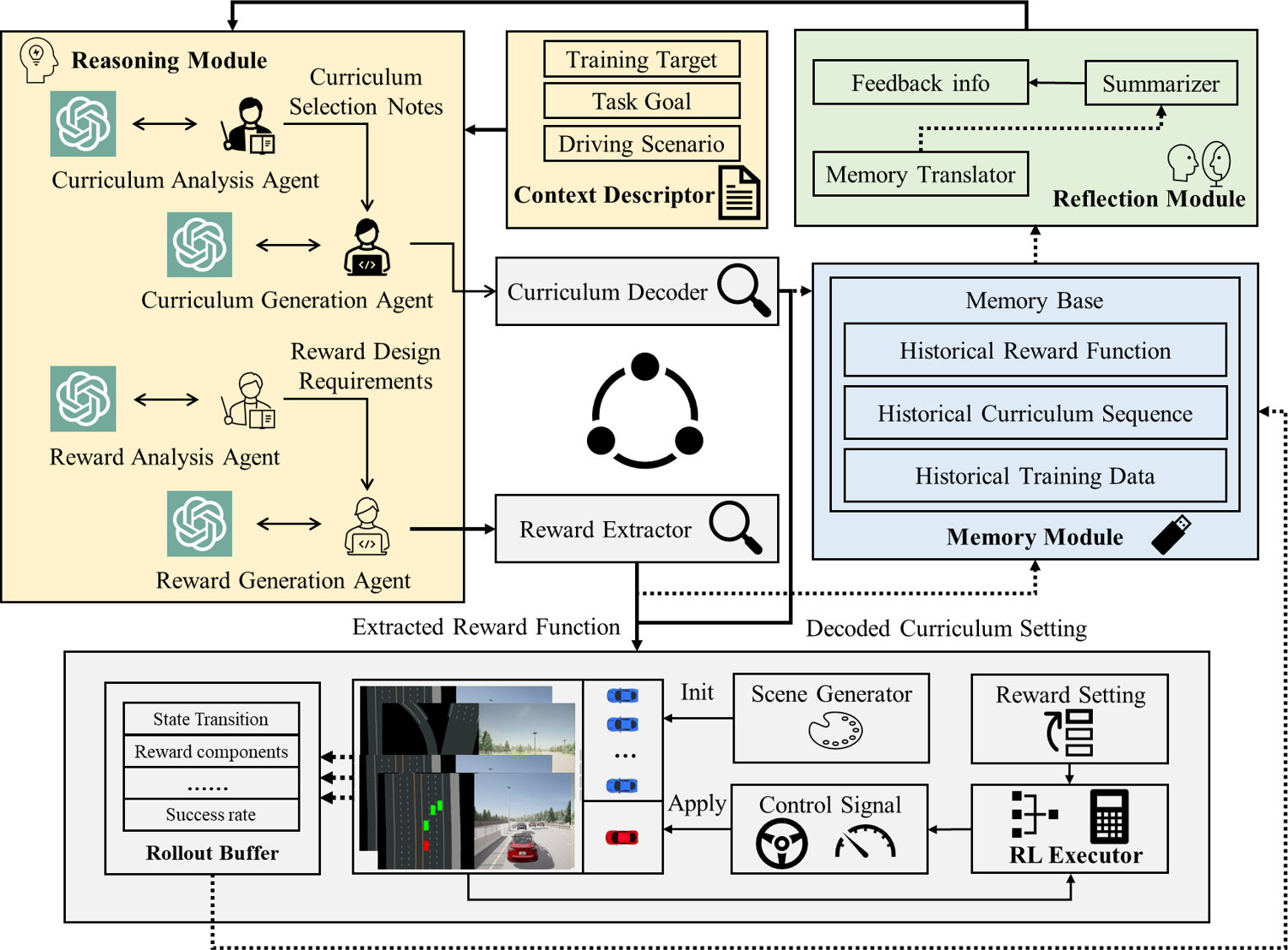}
   \caption{Overview of the LearningFlow framework for automated driving policy learning with interactive SVs. In the reasoning module, the analysis agents process prompts containing context descriptors, historical training information, and task objectives to perform inference, providing task analysis as a reference for the generation agent. The generation agent, based on the analysis results and other relevant prompts, selects training curricula and generates reward functions to initialize the downstream training environment and RL agent. After initialization, the RL agent interacts with the environment and records training data. Upon completing a certain number of episodes, the training data, along with the decision contents from the LLM agents, are stored in the memory module. These records are then summarized by the reflection module and fed back to the agents in the reasoning module to support the next round of inference.
   }
    \label{frame:learningflow}
\end{figure*}

\section{Methodology}

\subsection{Overview of the LearningFlow}

The detailed architecture of the proposed LearningFlow is illustrated in Fig. \ref{frame:learningflow}. 
First, general knowledge prompts related to autonomous driving tasks are generated separately to the curriculum analysis agent and reward analysis agent to analyze training tasks. Then the generated key points on curriculum generation and reward generation, along with general knowledge prompts for CL, reward generation, and code generation, are input into the curriculum generation agent and reward generation agent to facilitate training course selection and reward generation. Subsequently, the relevant training curricula and reward function codes are extracted from responses of the LLM to initialize the RL agent and interactive environment. Finally, the downstream RL executor explores and learns within the environment and reward functions designed by the LLM agents. 
The responses from LLM agents, along with training history, are recorded in a memory module; and after a specified number of episodes, they are fed back to the LLM agents to update the training curricula and reward functions.

In this work, the proposed LearningFlow framework consists of a reasoning module, a reflection module, and a memory module for curriculum generation and reward generation. By providing appropriate prompts to the proposed framework, the system is able to perform automated policy learning for various autonomous driving tasks. 
Details will be discussed in the following sections.

\subsection{Memory Module for Closed-Loop Policy Training Workflow}

LLMs inherently lack persistent memory capabilities, meaning that they cannot retain interaction information from previous sessions when processing new queries. Consequently, LLMs are unable to recall past exchanges with the user during subsequent reasoning tasks. 
Therefore, a memory module is introduced to store historical information from the training process to enable closed-loop online tuning workflow throughout the training process. This includes inference results from various LLM agents, generated training curricula, designed reward functions, recorded total reward and the individual reward components during training, and training metrics. 
The extracted information is stored in textual or vectorized form via an extraction module and integrated into prompts for the next session through a reflection module. By incorporating the memory module, LLMs can retain and leverage cross-session historical training data, enabling closed-loop reasoning and automated training adjustments, thereby significantly enhancing the efficiency and adaptability of policy learning.

\subsection{Iterative Curriculum Sequence Generation}

\subsubsection{Context Descriptor}

Given the generality of LLMs, it is necessary to provide relevant contextual information to help them understand task objectives. In this study, we use contextual descriptors to describe the current training task in natural language comprehensively. Considering the complexity of the task, the description includes characteristics of the driving scenario and objectives of policy learning. 
For the curriculum generation LLM agent, a carefully designed two-layer curriculum set and curriculum descriptions are included as part of the context. 
The generated description is then integrated into the system prompts of different agents to facilitate task-specific reasoning of curriculum analysis and generation.

\subsubsection{Curriculum Analysis Agent}

Our target scenarios are complex and challenging, involving not only SVs with diverse driving styles but also diverse traffic densities. To enhance sample efficiency, the designed two-layer curriculum set incorporates not only these factors but also subsets of curricula with different motion modalities of SVs. 
The unquantifiable nature of driving difficulty across different scenarios poses challenges to traditional curriculum selection algorithms and heightens the requirements for the curriculum generation agent.

Directly generating training curricula could result in LLMs producing low-quality outputs from the LLMs, such as inappropriate course transitions or the omission of critical curricula. To address this issue, we introduce a curriculum analysis agent. 
In the prompt for the curriculum analysis agent, we provide the foundational knowledge for CL, along with the structure of the two-layer curriculum set (\ref{2layer_curri_set}) and the design objectives for each layer. 
Specifically, the curriculum set is designed to guide the policy in progressively exploring the environment. It starts with simpler tasks and gradually transitions to more complex ones, thereby enhancing sample efficiency. 
Before generating training curricula, this agent incrementally analyzes the two-layer curriculum set by considering the current training context. 
It leverages contextual information from context descriptors, curriculum set information, historical curriculum sequence, current training progress, and historical training data of the RL policy. Based on this analysis, the agent identifies critical aspects to consider in the setup of the next round of CRL training and provides a rationale for the curriculum generation agent to make informed inferences.

\subsubsection{Curriculum Generation Agent}

After the curriculum analysis agent completes its reasoning, the analyzed results are passed to the curriculum generation agent as a reference. Specifically, the prompt generator utilizes the definition and analysis of the curriculum set, principles of CL, and output format prompts as textual context for training curriculum generation. It then directs the LLMs to perform reasoning and return the inferred results, which include the selected training curriculum. Then the selected curriculum is decoded from the response of the LLMs. Through the designed prompts and the collaboration with the curriculum analysis agent, the curriculum generation agent can effectively select appropriate curriculum tasks during the initialization phase and throughout the training process.

Given the complexity of environmental rules and state transitions, as well as the potential suboptimality in the selections generated by LLMs, we incorporate an $\epsilon$-greedy strategy, commonly used to address the exploration-exploitation dilemma in RL \cite{sutton2018reinforcement}, to configure the ultimate training environment. This strategy helps the RL agent better understand environmental features and adapt to environmental changes. The specific $\epsilon$-curriculum selection strategy is defined as follows:
\begin{equation}
C_k = \left\{
\begin{array}{l}
C_{LLM}, \ \text{with probability } 1 - \epsilon, \\
C_{random}, \ \text{with probability } \epsilon,
\end{array}
\right.
\label{equ:curr_gen}
\end{equation}
where $C_t$ is the employed curriculum setting at $k$-th episode, $C_{LLM}$, $C_{random}$ represent the curriculum selected by the agent and randomly, respectively; and $\epsilon$ is a parameter that decays over the training process.

\subsubsection{Curriculum Evaluation and Reflector}

In the initial phase of training, open-loop automatic curriculum generation is achieved through the collaboration of the analysis agent and the selection agent, which can provide RL agents in different training stages with suitable training environments. Meanwhile, interaction data generated during policy training is recorded in the memory module. To facilitate timely curriculum switching based on historical information, it is necessary to express in language whether the current curriculum is appropriate for subsequent training. For this purpose, a curriculum reflection module is introduced to generate feedback for the curriculum analysis and selection modules, enabling a closed-loop automatic curriculum agent workflow. 
Specifically, the curriculum reflection agent summarizes the characteristics of the historical curriculum sequence, the trajectory of policy rewards, and task performance metrics. It then generates reflection prompts to guide the analysis and selection of the next curriculum phase. 

The entire workflow of curriculum sequence generation can be expressed as follows: 
\begin{equation}
\left\{\begin{array}{l}
A_0^C = L(P_a^C), \\
C_0 = D_C(L(P_s^C,A_0^C)), \\
A_{n+1}^C=L(C_{hist}, A_{n}^C, P_a^C, P_f^C),\\
C_{n+1}=D_C(L(C_{hist}, P_s^C, P_f^C, A_{n+1}^C)) , \\
\hfill n=0,1,...,N_{max}.
\end{array}\right.
\label{equ:curriculum_agent}
\end{equation}
where $L$ refers to the LLM; \( A_n^C\), \(C_n \) represent the analysis and selected curriculum for the \( n \)-th training interval; \(C_{hist} \) is the historical curriculum sequence; \( P_a^C \), \( P_s^C \), and \( P_f^C \) denote the analysis prompt, selection prompt, and feedback prompt for the curriculum sequence generation, respectively; \(D_C(\cdot) \) is the decode function to extract the curriculum.

\subsection{Iterative Reward Generation} 

\subsubsection{Context Descriptor}
Similar to curriculum agents, it is essential to provide appropriate contextual information to guide LLMs in understanding the objectives of reward generation. Here, we also employ contextual descriptors to express the requirements of reward generation tasks in natural language comprehensively. Considering the intricacy of the reward design process, the descriptors include both general principles of reward functions and task-specific goals, such as the characteristics of driving interactions and the expected outcomes of RL agent training. These descriptions are incorporated into the system prompts of different agents to enable task-specific reasoning for reward analysis and generation.

\subsubsection{Reward Analysis Agent}

In our study, the complexity of autonomous driving tasks arises not only from varying driving scenarios but also from differing traffic densities. This increases the demand for the design of reward functions. 
Generating the reward function directly could lead to ineffective outputs from the LLMs. 

To address this issue, we introduce a reward function analysis agent that analyzes the task before the reward function is designed. 
For standard RL tasks, manually crafted reward functions typically incorporate accessible environment state variables, action variables, and constants. Therefore, it is reasonable to provide code segments with annotated notes containing accessible environmental variables as context for the reward analysis agent. This approach prevents exposing the internal mechanisms of the environment and eliminates the need for explicit state transition dynamics. 
The LLM agent leverages contextual information from context descriptors, the code segment of accessible variables, the next training environment determined by the curriculum agents, the current training progress, and the feedback data. This agent offers the key considerations during the reward generation process and provides a reference value range and analytical basis for the reward components, thus supporting the inference process of the reward generation agent.

\subsubsection{Reward Generation Agent} 

After the reward analysis agent completes its reasoning, the analyzed results and reference value range are passed to the reward generation agent. Besides, accessible variables, reward function signature, principles of reward function construction, and output format specifications are also provided as textual context for designing reward functions. 
Specifically, the generation of the reward function is required to output both the total reward and the individual components of the designed sub-rewards.
Given the above comprehensive information as context, a general reward generation prompt is crafted for input into the LLMs. 
Then the reward generation agent calls LLMs to perform reasoning to generate a response containing the designed reward function by utilizing its vast knowledge base and emergence ability. 
Subsequently, the executable code of the reward function is extracted from the response content and added to the environment program. Through carefully designed prompts and collaboration with the reward analysis agent, the reward generation agent can effectively construct appropriate reward functions throughout the training process.

\subsubsection{Reward Evaluation and Reflector}

Through the collaboration of multiple agents, an open-loop reward generation process is implemented to provide RL agents with an initial reward function. Then the RL agent can explore the environments with the guidance of the generated reward function and related data is collected for policy update and memory storage.

To facilitate the timely refinement of reward functions based on historical data, it is essential to evaluate and articulate whether the current reward function aligns with the training objectives. For this purpose, a reward reflection module is introduced to generate feedback prompts for the reward analysis and generation agents, forming a closed-loop reward generation process and online tuning workflow together. 
This reflection workflow generates reflection prompts to reshape and mutate the reward function based on feedback information and the previous reward function for subsequent training phases.

To enable effective closed-loop iteration, a comprehensive evaluation of the designed reward functions is required as a basis for subsequent improvements. It includes information from the interactions of the RL agent, such as historical data about the reward functions and their components, success rates, collision rates, and timeout rates. 
According to these metrics, the reward agents can integrate information at different levels of granularity. This includes coarse-grained data on the entire reward function and success rates, as well as fine-grained details on individual reward components. These capabilities allow for more effective optimization of the reward function design and the proposal of more targeted reward components. 
A prompt regarding context-based reward improvements is then provided to analysis and generation agents to propose a new, enhanced generation of the reward function based on previously designed ones. This prompt includes descriptions of the closed-loop feedback and suggestions for improvements, such as adding or removing reward components or modifying the reward coefficients.

The entire workflow of reward generation and online tuning can be expressed as follows: 
\begin{equation}
\left\{\begin{array}{l}
A_0^R = L(P_a^R), \\
R_0 = D_R(L(P_s^R,A_0^R)), \\
A_{n+1}^R=L(R_{hist}, A_{n}^R, P_a^R, P_f^R),\\
R_{n+1}=D_R(L(R_{hist}, P_s^R, P_f^R, A_{n+1}^R)) , \\
\hfill n=0,1,...,N_{max}.
\end{array}\right.
\label{equ:reward_agent}
\end{equation}
where \( A_n^R\), \(R_n \) represent the analysis and design reward for the \( n \)-th training interval; \(R_{hist} \) is the historical reward function; \( P_a^R \), \( P_s^R \), and \( P_f^R \) denote the analysis prompt, reward generation prompt, and feedback prompt for the reward function, respectively; \(D_R(\cdot) \) is the decode function to extract the reward function code.

\subsection{Downstream RL Executor}
\label{method:RL executor}

This work adopts an integrated decision-planning-control strategy using an RL policy as the downstream executor. Specifically, the RL policy generates decision variables based on observations, which are then used as the reference of the model predictive controller to calculate the control signal. The specific details are presented as follows. 
We use the notations of subscript $k$ to represent $k$-th time step within an episode. The observations of the RL agent are expressed as follows:
\begin{equation}
\begin{split}
\mathbf{O}_{k} &= \left[\ \mathbf{o}_{k}^0\ \ \mathbf{o}_{k}^1\ ...\ \mathbf{o}_{k}^{N_{\text{obs}}^{\max}}\ \right]^T \\
&= \begin{bmatrix}
  {\
\delta x_{k}^{0}}&{\delta y_{k}^{0}}&{v_{k}^{0}}&{\delta \psi_{k}^{0}} \\
  {\
\delta x_{k}^{1}}&{\delta y_{k}^{1}}&{\delta v_{k}^{1}}&{\delta \psi_{k}^{1}} \\
\vdots & \vdots & \vdots & \vdots \\
  {\
\delta x_{k}^{N_{\text{obs}}^{\max}}}&{\delta y_{k}^{N_{\text{obs}}^{\max}}}&{\delta v_{k}^{N_{\text{obs}}^{\max}}}&{\delta \psi_{k}^{N_{\text{obs}}^{\max}}} \\
\end{bmatrix},
\end{split}
\label{RL_obs_matrix}
\end{equation}
where $N_{\text{obs}}^{\max}$ is the maximum number of the observed SVs for the EV; $v^0_k$ denotes the current speed of the EV; $\delta x_{k}^{i}, \delta y_{k}^{i},\delta v_{k}^{i},$ and $\delta \psi_{k}^{i}$ represent the differences in the X-axis coordinate, Y-axis coordinate, speed, and heading angle, respectively, between the EV and the destination position ($i=0$) as well as the $i$-th SV ($i=1,2,\dots,N_{\text{obs}}^{\max}$). 

Here, the RL policy is represented by a neural network $\pi$ parameterized by $\bm{\theta}$. Given the RL observation $\mathbf{O}_k$ at time step $k$, the action of RL agent is generated by:
\begin{equation}
    a_k^{RL} = \pi_{\bm{\theta}}(\mathbf{O}_k)
    \label{eq:get_action}
\end{equation}

The specific definition of the sub-action spaces in (\ref{Action_space}) are introduced as follows. 
The waypoint sub-action space is defined as:
\begin{equation}
    A_1 = \left\{ \textup{WP}_0, \textup{WP}_1, ..., \textup{WP}_4\right\}, 
    \label{AS1}
\end{equation}
where $\textup{WP}_i= [x^{\text{WP}}_i\ y^{\text{WP}}_i\ \psi^{\text{WP}}_i]^T$ is the $i$-th waypoint, which includes the reference information about the X-axis, Y-axis coordinates, and heading angle of the waypoint, respectively. 
Waypoints are provided by a predefined road map and several path-searching methods, such as $A^*$ search algorithm. A reference waypoint set is generated at the beginning of the task. The 5 waypoints closest to the EV (${\textup{WP}^{\prime}_{i},i=0,1,...,4}$) are added to the $A_1$. 
The reference velocity sub-action space is defined as:
\begin{equation}
    A_2 = \left\{ 0,\frac{v_{limit}}{4},\frac{v_{limit}}{2},\frac{3v_{limit}}{4},v_{limit}\right\}, 
    \label{AS2}
\end{equation}
where $v_{limit}$ is speed limitation of the road. The lane change sub-action space is defined as:
\begin{equation}
    A_3 = \left\{ -1,0,1 \right\},
    \label{AS3}
\end{equation}
where $-1,0$, and $1$ represent left lane change, lane keeping, and right lane change maneuvers, respectively. 

The coordination of the above three sub-action spaces can enable flexible motion patterns for the EV to interact with SVs exhibiting diverse behaviors. When rapid movement toward the target is needed, the RL agent can select a distant waypoint and high speed; while it can choose the nearest waypoint and low speed for emergency braking. The lane-changing sub-action adds further flexibility. Selecting a closer waypoint on an adjacent lane indicates an urgent lane change, while a distant waypoint enables a smoother lane change for collision avoidance or overtaking. 
The reward function and the training environment are provided by the LLM agents. 
After the RL policy generates action outputs based on observations, these actions are decoded and passed to the model predictive controller, which converts them into execution trajectories and control commands to be applied to the EV.

Once the training curriculum and reward function are decoded from the answers of LLM agents, the RL agent explores the environment set by the configuration of the selected curriculum. 
Relevant historical training information is recorded in the replay buffer. 
Once a certain number of episodes have been gathered, the RL policy undergoes training to optimize the cumulative objective function that is associated with the sequence of generated training curricula $(C_0,C_1,...,C_{N_{max}+1})$ as follows:
\begin{equation}
\bm{\theta}^*=\arg \max _{\bm{\theta},(C_0,C_1,...,C_{N_{max}+1})} J(\bm{\theta}),
\label{equ:ACRL}
\end{equation}
where $J(\bm{\theta})$ denotes the objective function for the RL policy with parameter $\bm{\theta}$.

\begin{figure}[t] % [t]
\centering
    \subfigure[Curriculum analysis agent]{\includegraphics[width=.45\textwidth]{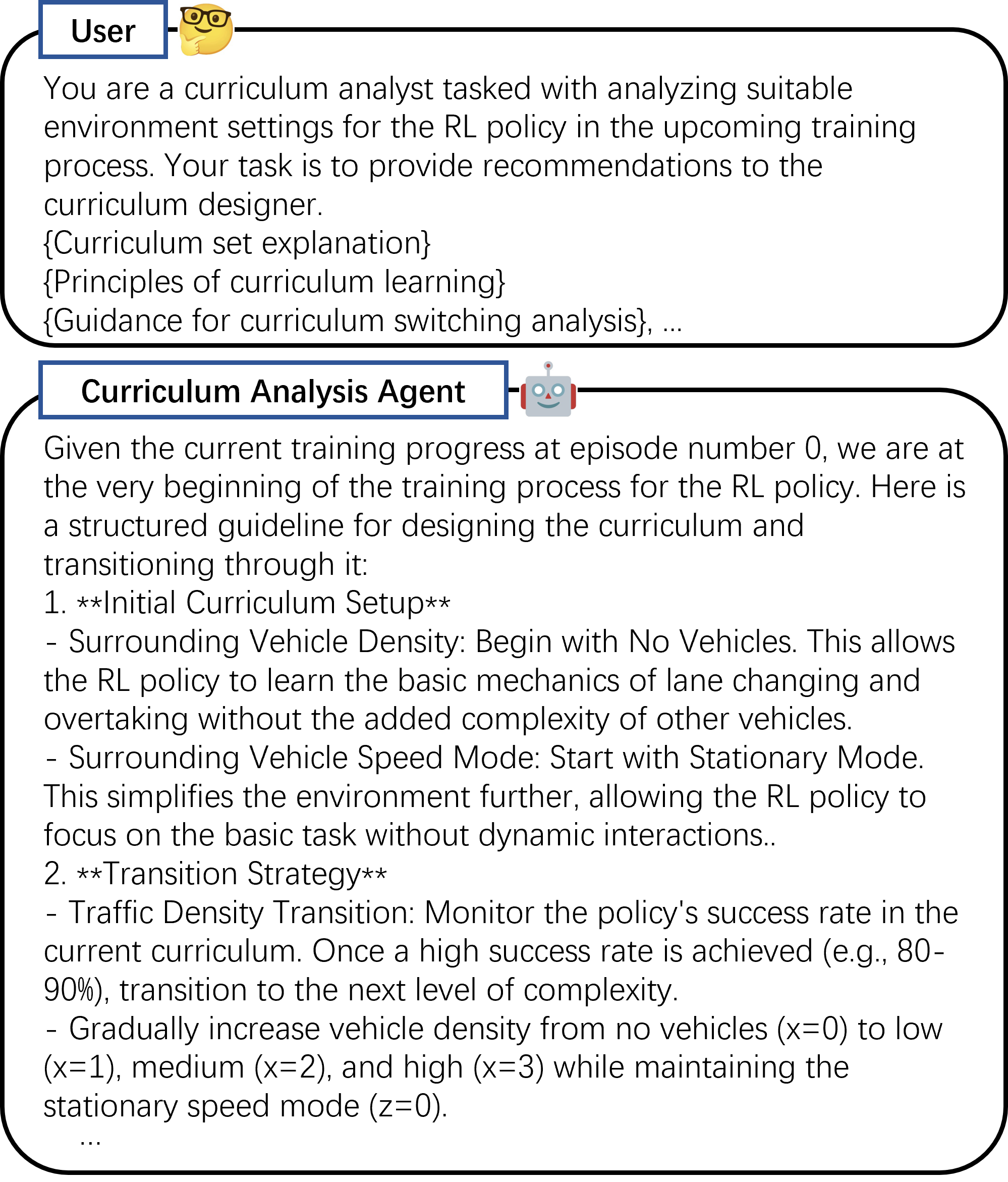}\label{exp:caa}}
    \subfigure[Curriculum generation agent]{\includegraphics[width=.45\textwidth]{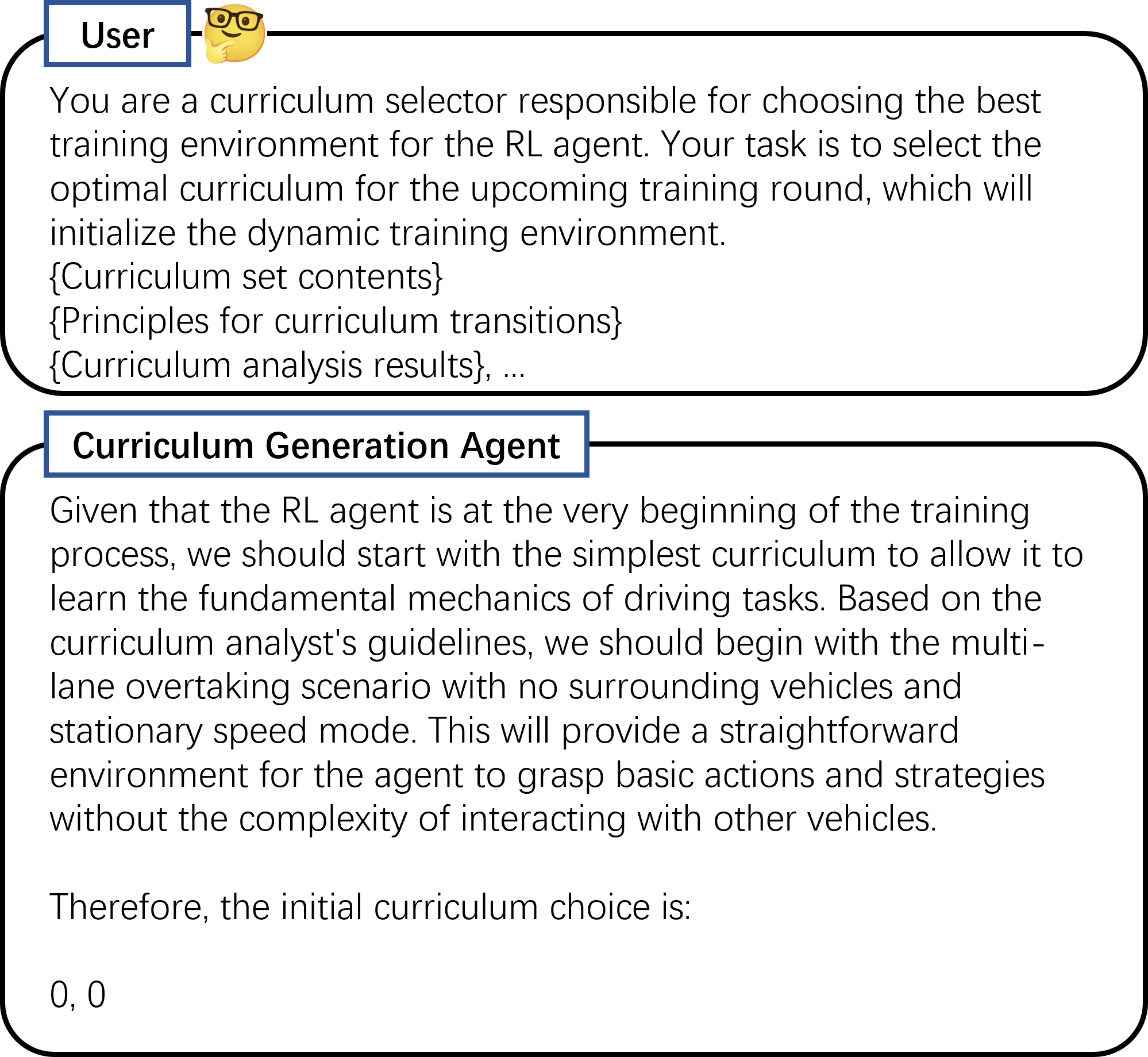}\label{exp:cga}}
\caption{Representative segments of the curriculum generation process demonstration during the initial training phase.
}
\label{exp:curri}
\end{figure}

\begin{figure}[!htbp] % [t]
\centering
    \subfigure[Reward analysis agent]{\includegraphics[width=.45\textwidth]{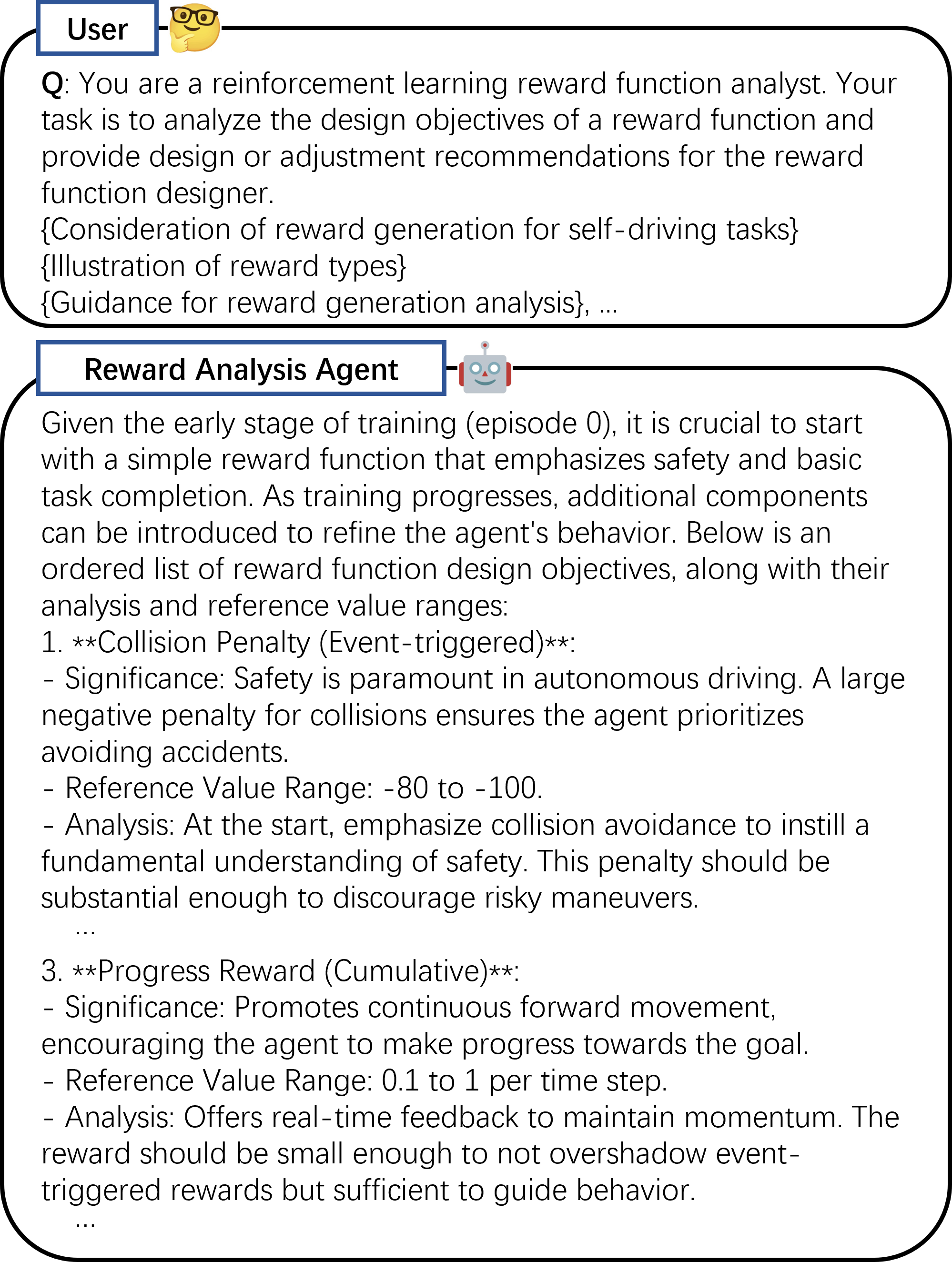}\label{exp:raa}}
    \subfigure[Reward generation agent]{\includegraphics[width=.45\textwidth]{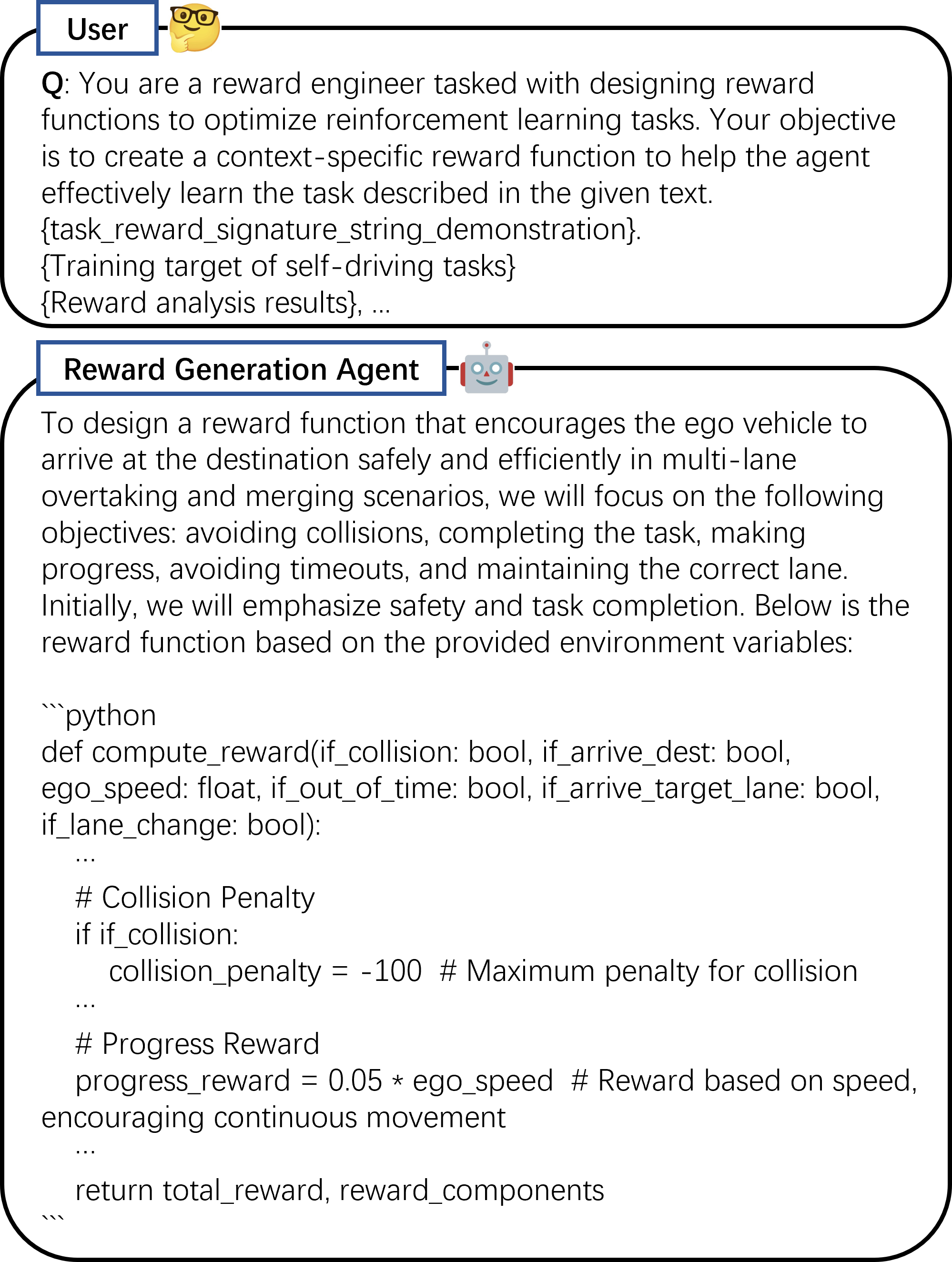}\label{exp:rga}}
\caption{Representative segments of the reward generation process demonstration during the initial training phase.
}
\label{exp:reward}
\end{figure}

\section{Experiments}

\subsection{Experimental Setup}

In this section, we implement the automated policy learning workflow in two urban driving scenarios. 
The experiments are carried out on the Ubuntu 18.04 system with Intel(R) Core(TM) i9-14900K CPU and NVIDIA GeForce RTX 3090 GPU. 
All self-driving scenarios involved in experiments are constructed on the CARLA simulator \cite{dosovitskiy2017carla}. 
Here, we select the Tesla Model 3 as the EV and SVs. 
To validate the capability of LearningFlow in automating policy learning across different driving tasks, an on-ramp merging scenario and a multi-lane overtaking scenario from the Town06 map are chosen for validation. These two scenarios are essential and classic components of urban driving, each with distinct operational characteristics. The overtaking task requires the EV to engage in multiple rounds of interactive decision making with SVs moving in parallel, ensuring a safe overtaking maneuver without disrupting other vehicles. In contrast, the merging task necessitates an accurate assessment of the inter-vehicle gap sizes and driving intentions of SVs on the main road, allowing the EV to enter from a narrow ramp within a limited time and distance. Failure to merge within this time window could force the vehicle to decelerate suddenly or even stop at the end of the lane, which complicates the decision-making process.

Since the task involves selecting curricula and designing reward functions based on contextual information from multiple information sources, robust reasoning capabilities of LLMs are necessary. Additionally, unlike tasks in the LLMs-in-the-task-loop approaches, our LearningFlow framework only calls the LLMs during the curriculum generation and reward design processes, meaning that the frequency of LLM calls is relatively low. Taking this into consideration, we choose the GPT-4o model for our experiments, which strikes a balance between reasoning capability and performance requirements. 

Specifically, proximal policy optimization (PPO) \cite{schulman2017proximal} is chosen as the RL policy training algorithm considering its ability to balance training stability and efficiency in data utilization. This feature makes it particularly suitable for the stated driving tasks that are investigated.  Additionally, PPO is well-suited for multi-dimensional discrete action spaces, which is essential in this work. The clipped objective function of the PPO algorithm is utilized to update the RL policy, which is shown as follows:
\begin{equation}
J_k(\bm{\theta})=\mathbb{E}_k\left[\min \left(\rho_k(\bm{\theta}) \hat{A}_k, \operatorname{clip}\left(\rho_k(\bm{\theta}), 1-\epsilon, 1+\epsilon\right) \hat{A}_k\right)\right],
\end{equation}
where $\rho_k(\bm{\theta})$ represents the probability ratio of the new policy to the old policy; $\hat{A}_k$ denotes the estimator of the advantage function at time step $k$; $\epsilon$ is the clip parameter. 

In this work, the actor-critic architecture is adopted to implement the PPO policy of the RL executor. 
The action network and critic network are set as fully connected networks with 2 hidden layers of 256 units and 128 units by PyTorch and trained with the Adam optimizer. 
The number of epochs is set to 50. The learning rate of the action network and critic network are set to $5 \times 10^{-4}$ and $1 \times 10^{-3}$, respectively. $\gamma$ is set to 0.99. 
$\eta$ is set to 0.2. The update frequency parameters $N_{P}, N_C$, and $N_R$ are set to 50, 100, and 1000, respectively. 
The low-level model predictive control optimization problem is solved by CasADi \cite{andersson2019casadi}, with the IPOPT option and single-shooting approach. 
We train the RL policy using the proposed LearningFlow method within the experimental scenarios, where modifications are made only to the textual descriptions of task characteristics across different scenarios.

In this work, the proposed framework is compared with the following approaches as follows: 

\begin{itemize}
    \item Vanilla PPO: the vanilla PPO policy is directly trained by PPO algorithm \cite{schulman2017proximal} in the task scenario. 
    \item AutoReward: a state-of-the-art approach which iteratively refines the reward function generated by LLMs after the whole training \cite{han2024autoreward}. The policy is directly trained in the task scenario. 
    Only the code of the environmental observation part is provided, which is the same as LearningFlow. 
    \item LearningFlow without analysis process (w/o AP): the proposed method without the analysis agents. 
\end{itemize}

For the sake of fairness, the parameters of the downstream RL executors for all methods are set to be identical. The clip parameter is configured to 0.2. The SVs operate in an autopilot mode provided by the CARLA simulator, with their driving styles being randomly assigned.

The traffic density settings in the scenarios are introduced as follows. Empty indicates that no SVs are present. In the on-ramp merging scenario, low, medium, and high densities correspond to 2, 4, and 8 SVs, respectively, in the two nearest lanes of the main road near the merging lane. In the multi-lane overtaking scenario, these densities represent the presence of 1, 2, and 3 SVs in front of the EV. These settings introduce a variety of traffic conditions and potentially require the EV to interact with different numbers of SVs, thereby increasing the complexity of the driving tasks. 
We begin by training the RL policy in multi-lane overtaking and on-ramp merging scenarios, respectively. Subsequently, we evaluate the performance of the trained RL policies in these scenarios, testing them with diverse traffic densities, and across different driving tasks.

\begin{figure}[t] %[htbp]
    \centering
    \includegraphics[trim=0 0cm 0 0.5cm, width=0.9\linewidth]{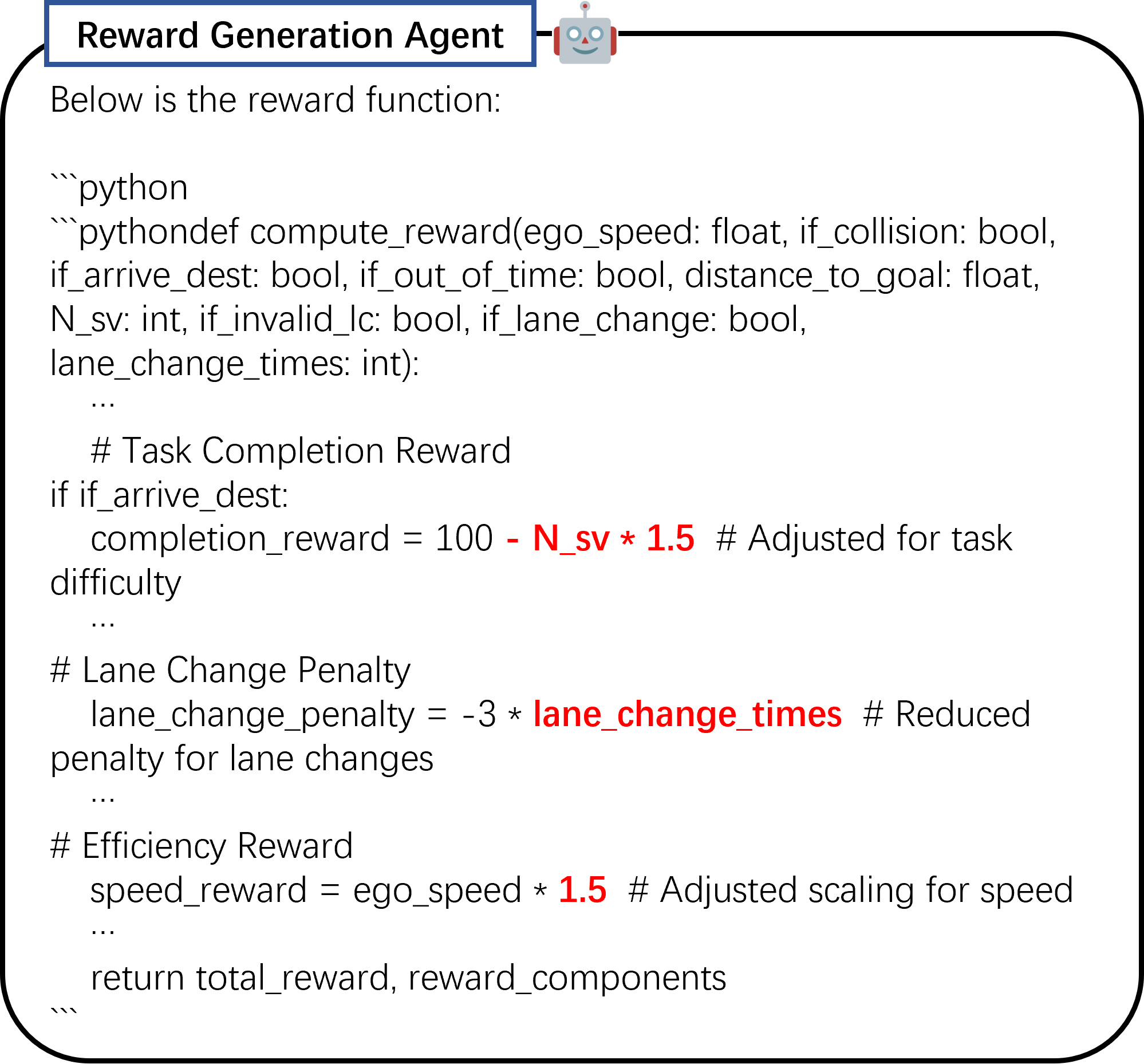}
   \caption{A failure case of reward generation process by LearningFlow without analysis agents, where the design flaws are highlighted in bold red.
   }
    \label{exp:failure_case}
\end{figure}

\begin{table*}[!htbp]
\renewcommand{\arraystretch}{1} 
\scriptsize
\centering
\caption{Performance comparison among different methods and different driving tasks.
}
\label{table:test_res}
\resizebox{\linewidth}{!}{
\begin{tabular}{cl|ccc|ccc|ccc|ccc}
\hline
\multicolumn{2}{c|}{\multirow{2}{*}{Methods}}                             & \multicolumn{3}{c|}{Empty}             & \multicolumn{3}{c|}{Low Density}      & \multicolumn{3}{c|}{Medium Density}    & \multicolumn{3}{c}{High Density}       \\ \cline{3-14} 
\multicolumn{2}{c|}{}                                                     & S(\%)        & C(\%)      & TO(\%)     & S(\%)       & C(\%)      & TO(\%)     & S(\%)       & C(\%)       & TO(\%)     & S(\%)       & C(\%)       & TO(\%)     \\ \hline
\multicolumn{1}{c|}{\multirow{2}{*}{Vanilla PPO}}            & Overtaking & 99           & 1          & 0          & 80          & 20         & 0          & 69          & 31          & 0          & 62          & 38          & 0          \\
\multicolumn{1}{c|}{}                                        & Merging    & 100          & 0          & 0          & 89          & 11         & 0          & 75          & 25          & 0          & 68          & 32          & 0          \\ \hline
\multicolumn{1}{c|}{\multirow{2}{*}{AutoReward (iter=0)}}    & Overtaking & 99           & 0          & 1          & 73          & 27         & 0          & 60          & 40          & 0          & 48          & 52          & 0          \\
\multicolumn{1}{c|}{}                                        & Merging    & 0            & 100        & 0          & 82          & 18         & 0          & 71          & 29          & 0          & 51          & 49          & 0          \\ \hline
\multicolumn{1}{c|}{\multirow{2}{*}{AutoReward (iter=5)}}    & Overtaking & 100          & 0          & 0          & 85          & 15         & 0          & 76          & 24          & 0          & 70          & 30          & 0          \\
\multicolumn{1}{c|}{}                                        & Merging    & 100          & 0          & 0          & 94          & 6          & 0          & 85          & 15          & 0          & 71          & 29          & 0          \\ \hline
\multicolumn{1}{c|}{\multirow{2}{*}{LearningFlow (w/o AP)}}  &        Overtaking & 100          & 0          & 0          & 89          & 11         & 0          & 75          & 25          & 0          & 69          & 31          & 0   \\
\multicolumn{1}{c|}{}                                        & Merging    & 100          & 0          & 0          & 91          & 9          & 0          & 83          & 17          & 0          & 72          & 28          & 0          \\ \hline
\multicolumn{1}{c|}{\multirow{2}{*}{\textbf{LearningFlow}}} & Overtaking & \textbf{100} & \textbf{0} & \textbf{0} & \textbf{97} & \textbf{3} & \textbf{0} & \textbf{90} & \textbf{10} & \textbf{0} & \textbf{85} & \textbf{15} & \textbf{0} \\
\multicolumn{1}{c|}{}                                        & Merging    & \textbf{100} & \textbf{0} & \textbf{0} & \textbf{99} & \textbf{1} & \textbf{0} & \textbf{94} & \textbf{6}  & \textbf{0} & \textbf{87} & \textbf{13} & \textbf{0} \\ \hline
\end{tabular}
}
\begin{tablenotes}[flushleft] 
\item \textit{Note:} S, C, and TO represent success rate, collision rate, and timeout rate, respectively.
\end{tablenotes}
\end{table*}

\subsection{Demonstration of Collaboration Among Multiple LLM Agents}

To illustrate the collaboration among multiple LLM agents during the policy training process, we selected an example of analysis and content generation from the cooperative interactions of four LLM agents, as shown in Figs. \ref{exp:curri}-\ref{exp:reward}. As a comparison, a reward generation demonstration of LearningFlow without the analysis process is shown in Fig. \ref{exp:failure_case}.

As shown in Fig. \ref{exp:curri}, at the beginning of training, the curriculum analysis agent evaluates the upcoming training curriculum based on the training progress and the core principles of CL, considering factors such as initial curriculum settings, switching strategies, and switching criteria. Subsequently, the curriculum generation agent selects a suitable training curriculum from the curriculum set by leveraging the analyzed results and adhering to CL principles, generating the selection in a standard format. During the subsequent training process, the curriculum analysis and generation agents iteratively analyze and evaluate training curricula by incorporating the historical curriculum sequence and RL policy training data, enabling automatic curriculum switching and thereby improving the sample efficiency of policy learning.

In Fig. \ref{exp:reward}, during the early stages of training, the reward analysis agent begins by examining the current performance of the policy, considering various factors relevant to the autonomous driving task, and analyzing each reward component, its significance, and the reference value range that should be considered in the reward generation process. Next, the reward generation agent, based on this analysis, generates a reward function in a standardized format, adhering to the function's signature and coding conventions. The generated reward function is then extracted and embedded into the environment code for training. After a certain number of episodes, the reward analysis and generation agents adjust the reward components and coefficients based on the policy's performance, providing reasons for the adjustments along with the reference value range. Through the collaboration of the reward function agents, we achieve the automatic design and online adjustment of the reward function, thereby enhancing the performance of the policy.

In the failure case shown in Fig. \ref{exp:failure_case}, the reward generation agent fails to generate an appropriate reward function. Specifically, the reward generation agent misinterprets the expected impact of traffic density $N\_sv$ on the reward function, reversing the operator, which hinders the RL policy when exploring tasks with higher traffic density. Furthermore, without the support of analysis agents, the reward generation agent fails to accurately understand the contribution of the $lane\_change\_times$ variable in the lane change penalty, which should be calculated in an event-triggered manner. Additionally, it also leads to difficulties in designing reasonable coefficient ranges. For instance, the coefficient and calculation method of the speed reward term could cause its accumulated value to exceed the reward for task completion, leading the agent to prioritize speed rewards at the expense of driving safety and encouraging reckless behavior. This demonstration highlights the crucial role of the analysis agent in the LearningFlow framework.

\begin{figure*}[!htbp] % [t]
\centering
    \subfigure[Multi-lane overtaking task. The EV is initialized in the middle lane, and its target position is set to the end of this road.]{\includegraphics[width=0.95\textwidth]{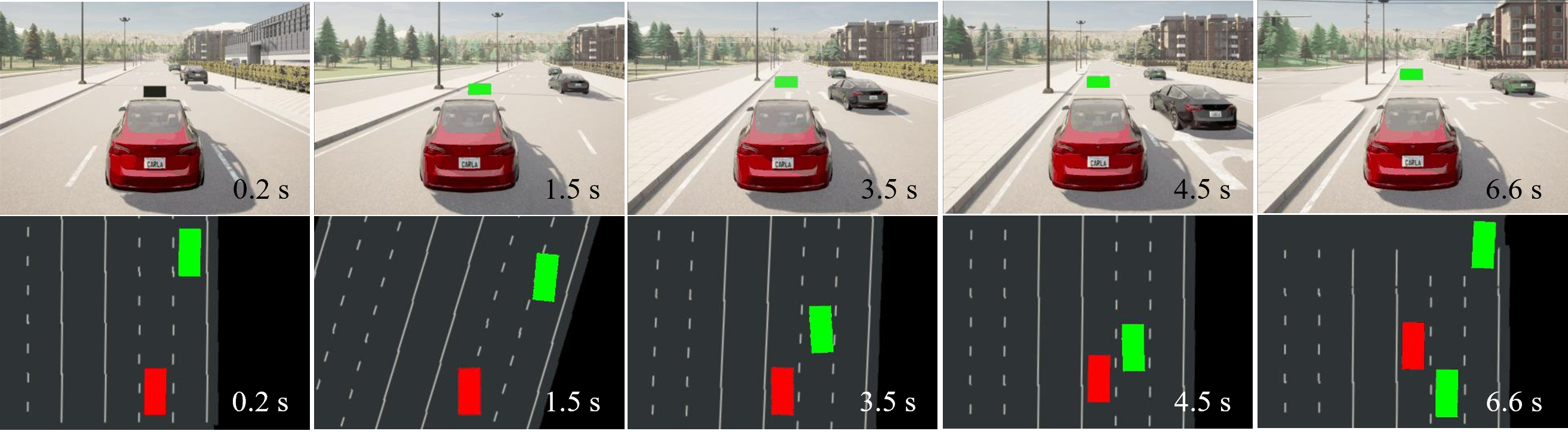}\label{exp:demo_overtaking}}
    \subfigure[On-ramp merging task. The EV is initialized in the merging lane, and its target lane is set to the second lane counted from left to right on the main road.]{\includegraphics[width=0.95\textwidth]{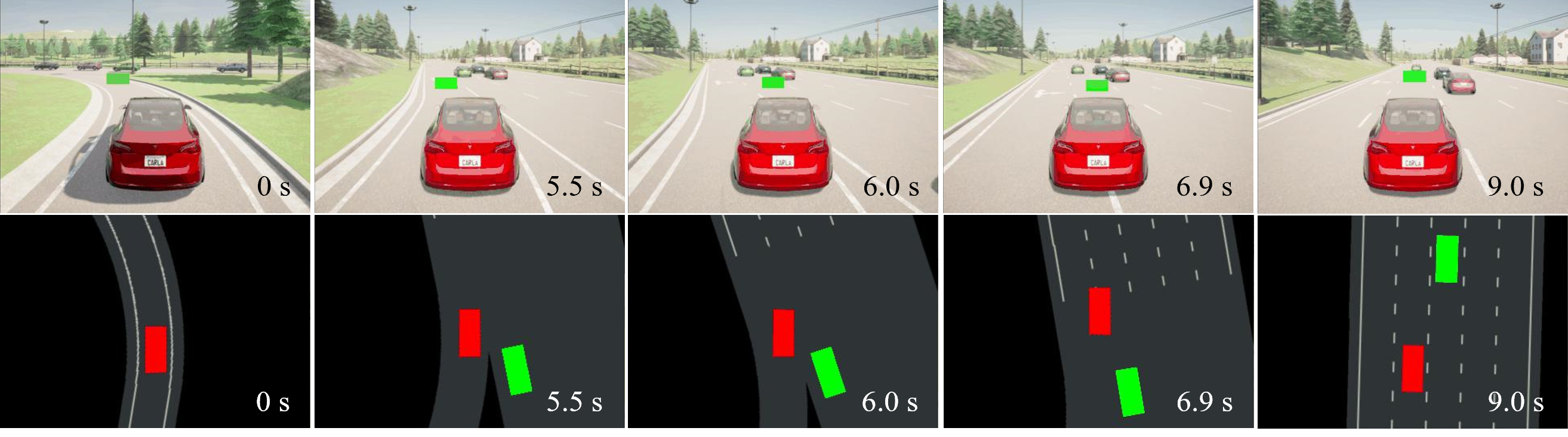}\label{exp:demo_merging}}
\caption{Key frames of two demonstrations using our method in multi-lane overtaking and on-ramp merging scenarios within CARLA. The upper and lower parts of the sub-figures show third-person views of the EV and bird-eye views, respectively. In the third-person sub-figures, the green rectangles represent the intermediate points determined by the RL policy. In the bird-eye view sub-figures, the red rectangle corresponds to the EV, and the green rectangles denote the SVs. 
}
\label{fig:demo_2tasks}
\end{figure*}

\subsection{Comparative Results and Analysis}

To quantitatively compare the performance of LearningFlow with the baseline methods, we conduct statistical testing on trained policies trained by all approaches in the on-ramp merging scenario and the multi-lane-overtaking scenario with the different traffic densities. Each learned policy undergoes 100 repeated tests in each task setting. 
The results are summarized in Table \ref{table:test_res}.

The table shows that the proposed approach achieves the highest success rate across all testing driving tasks and traffic configurations. Although the success rate tends to decrease as the number of SVs increases, this outcome demonstrates the effectiveness of the LearningFlow framework in the automated policy learning process. It can help to obtain a more effective policy within the same number of training episodes, thus improving the sample efficiency. 
Overall, this result demonstrates the effectiveness and generalization capability of LearningFlow across various driving tasks.

\begin{table*}[!htbp]
\renewcommand{\arraystretch}{1}
\scriptsize
\centering
\caption{Performance comparison at the multi-lane overtaking scenario among different RL algorithms.
}
\label{table:test_res_rl}
\resizebox{\linewidth}{!}{
\begin{tabular}{c|ccc|ccc|ccc|ccc}
\hline
\multirow{2}{*}{Methods} & \multicolumn{3}{c|}{Empty} & \multicolumn{3}{c|}{Low Density} & \multicolumn{3}{c|}{Medium Density} & \multicolumn{3}{c}{High Density} \\ \cline{2-13} 
                         & S(\%)   & C(\%)  & TO(\%)  & S(\%)     & C(\%)    & TO(\%)    & S(\%)      & C(\%)     & TO(\%)     & S(\%)     & C(\%)    & TO(\%)    \\ \hline
Vanilla DQN              & 100     & 0      & 0       & 78        & 22       & 0         & 70         & 30        & 0          & 59        & 41       & 0         \\ \hline
LearningFlow DQN         & 100     & 0      & 0       & 92        & 8        & 0         & 86         & 14        & 0          & 81        & 19       & 0         \\ \hline
Vanilla SAC              & 100     & 0      & 0       & 82        & 18       & 0         & 72         & 28        & 0          & 63        & 37       & 0         \\ \hline
LearningFlow SAC         & 100     & 0      & 0       & 93        & 7        & 0         & 83         & 17        & 0          & 79        & 21       & 0         \\ \hline
\end{tabular}
}
\begin{tablenotes}[flushleft] % 使用flushleft选项来实现左对齐
\item \textit{Note:} S, C, and TO represent success rate, collision rate, and timeout rate, respectively.
\end{tablenotes}
\end{table*}

Additionally, it is observed that methods utilizing LLMs to assist the training of the RL policy generally outperform the vanilla PPO in most of the test tasks. It is noted that the RL policy trained by the AutoReward method with zero iterations fails to complete the merging task without SVs, likely due to the suboptimal design of the reward function, which arises from the lack of access to the internal environment code. However, this issue is resolved after the reward function undergoes iteration, as demonstrated by AutoReward (iter=5). Moreover, a comparison between the results of LearningFlow and LearningFlow (w/o AP) reveals that the absence of the analysis agent leads to a decline in policy performance. This could be attributed to the tendency of the generation agent to produce erroneous results when dealing with complex tasks, resulting in training instability. This highlights the crucial role of the analysis agent in enhancing the effectiveness of policy training.

\subsection{Demonstration of Maneuvering Abilities in Driving Tasks} 

To illustrate the actual overtaking and merging capabilities, we select one outcome from the testing results of all driving tasks using the proposed approach. The snapshots of these two examples are shown in Fig. \ref{fig:demo_2tasks}.

\begin{itemize}
    \item \textbf{Multi-lane overtaking task}: In this example, the EV is initialized in the middle lane, while two SVs are initialized at different positions in the right lane. At $0.2$ s, the RL policy outputs a reference point ahead, guiding the EV to accelerate for overtaking. Then, at $1.5$ s, the RL policy detects a nearby SV performing a left lane change action. In response, the policy outputs a reference point to the left ahead, guiding the EV to perform the overtaking lane change, which is completed in $3.5$ s, followed by acceleration. Finally, between $4.6$ s and $6.6$ s, the EV successfully completes the overtaking maneuver and reaches the target area under the guidance of the RL policy, completing the multi-lane overtaking task safely and efficiently.
    \item \textbf{On-ramp merging task}: At the beginning of the demonstration, the EV is initialized at a random position on the merging lane. At this point, the RL policy outputs a reference point ahead, guiding the EV towards the main road. Then, at $5.5$ s, the EV enters the merging zone. The RL policy detects an SV approaching from the rear-right and outputs a reference point ahead to accelerate and increase the distance. At $6.0$ s, the rear-right SV slows down and yields, maintaining a safe distance from the EV. The RL policy then outputs a reference point in the right lane ahead, guiding the EV to merge into the target lane. Finally, between $6.9$ s and $9$ s, the RL policy selects a distant reference point and sets a high reference speed, guiding the EV to smoothly complete the merging maneuver and reach the target area. Ultimately, under the guidance of the RL policy, the EV successfully and efficiently completes the on-ramp merging task.
\end{itemize}

\subsection{Adaptation to Different RL Algorithms} 

To evaluate the generalization of the proposed LearningFlow framework across different RL algorithms, we replace the downstream PPO algorithm with DQN and SAC algorithms and conducted training and testing on the multi-lane overtaking task. The test results are shown in Table \ref{table:test_res_rl}. 
As seen from the statistics in the table, LearningFlow effectively facilitates policy learning across different RL algorithms, improving the success rate of the trained policies in task scenarios. Compared to the RL policies trained with the PPO algorithm in Table \ref{table:test_res}, the success rates for DQN and SAC are slightly lower. This can be attributed to the fact that the performance of the RL policies also depends on the configuration of the action space and state space. Specifically, DQN could not perform as well as PPO in handling continuous state spaces and multi-dimensional discrete action spaces. Meanwhile, SAC, which is designed for continuous action spaces, is inherently limited in tasks involving multi-dimensional discrete action spaces. Nevertheless, in general, the RL policies trained with the LearningFlow framework show enhanced performance compared to Vanilla RL algorithms. This demonstrates that the proposed approach for reward and curriculum generation is adaptable to various RL algorithms, and this showcases the generalization capability of the proposed framework.

\vspace{-0.1cm}

\section{Conclusion}

In this paper, we propose LearningFlow, a closed-loop automated policy learning workflow for autonomous driving, which utilizes the collaboration of LLM agents to generate and dynamically adjust the training curriculum and reward functions of the RL policy. By incorporating analysis and generation agents, our approach enhances the understanding of complex driving tasks, thereby improving the efficiency of automatic curriculum and reward generation, as well as the performance of RL policies. Experimental results demonstrate the effectiveness of the proposed framework. Compared to baseline and SOTA methods, the RL policy trained by LearningFlow achieves the highest success rate. 
Furthermore, ablation studies validate the effectiveness of our policy learning workflow. 
Demonstrations in various driving scenarios, involving interactions with SVs, showcase the superior interaction-awareness capability of our method. Finally, experimental results indicate that the proposed method is highly compatible with various RL algorithms. 
Future work includes incorporating advanced diffusion models to enhance the multimodal decision-making capability of the downstream RL executor, thereby further improving the generalization ability of RL solutions.

\vspace{-0.1cm}

\bibliographystyle{IEEEtran}
\bibliography{pzqbib}

\vfill

\end{document}